\documentclass{svproc}
\usepackage{url}
\usepackage[sorting=none, style=nature]{biblatex}
\usepackage{graphics}
\usepackage{caption}
\usepackage{subcaption}
\usepackage{amsmath}
\usepackage{amssymb}
\usepackage{lipsum}
\usepackage{graphicx}
\usepackage{algorithm}
\usepackage{algpseudocode}
\usepackage{romannum}
\usepackage{physics}
\usepackage{hyperref}
\usepackage{enumitem}
\usepackage{siunitx}
\newlength{\twosubht}
\newsavebox{\twosubbox}

\begin{document}
\mainmatter              
\title{Soft Gripping System for Space Exploration Legged Robots}
\titlerunning{Soft Gripping System for Space Exploration}  
%
\author{Arthur Candalot\inst{1} \and Malik-Manel Hashim\inst{1}, Brigid Hickey\inst{1}, Mickael Laine\inst{1} \and Mitch Hunter-Scullion\inst{2} \and Kazuya Yoshida\inst{1}}
\authorrunning{A. Candalot et al.} 
%
\tocauthor{Arthur Candalot, Malik Manel Hashim, Brigid Hickey, Mickael Laine, Mitch Hunter-Scullion and Kazuya Yoshida}
\institute{Space Robotics Lab., Tohoku University, Sendai 980-8579, Japan,\\
\email{acandalot@dc.tohoku.ac.jp}, 
\and
Asteroid Mining Corporation Ltd., London E14 9HQ, United Kingdom.
}

\maketitle              

\begin{abstract}
Although wheeled robots have been predominant for planetary exploration, their geometry limits their capabilities when traveling over steep slopes, through rocky terrains, and in microgravity. Legged robots equipped with grippers are a viable alternative to overcome these obstacles. This paper proposes a gripping system that can provide legged space-explorer robots a reliable anchor on uneven rocky terrain. This gripper provides the benefits of soft gripping technology by using segmented tendon-driven fingers to adapt to the target shape, and creates a strong adhesion to rocky surfaces with the help of microspines. The gripping performances are showcased, and multiple experiments demonstrate the impact of the pulling angle, target shape, spine configuration, and actuation power on the performances. The results show that the proposed gripper can be a suitable solution for advanced space exploration, including climbing, lunar caves, or exploration of the surface of asteroids. 
\keywords{Soft gripper, Climbing robots, Asteroid exploration, \\Microspine gripper}
\end{abstract}
\section{Introduction}
\label{introduction}

The discovery and exploitation of critical resources present in space is now being proposed as one of the most viable solutions to facilitate future space development \cite{asteroidmining}. To do so, the development of highly capable autonomous robots to navigate and operate freely on planetary and asteroid surfaces is required \cite{multileg}. Hexapods, and more generally walking robots, will be necessary as they offer stable locomotion, leg versatility, and endurance \cite{hexapod} to perform specific tasks in space. 

Thanks to the success of the Hayabusa missions by JAXA, the surface of the asteroids Itokawa and Ryugu (Fig.~\ref{fig:ryugu}) are now known to be composed of many random-shaped and extremely uneven rocks. 

\subsection{SCAR-E robot}

\begin{figure}[t]
    \sbox\twosubbox{%
        \resizebox{\dimexpr.9\textwidth-1em}{!}{%
            \includegraphics[height=3cm]{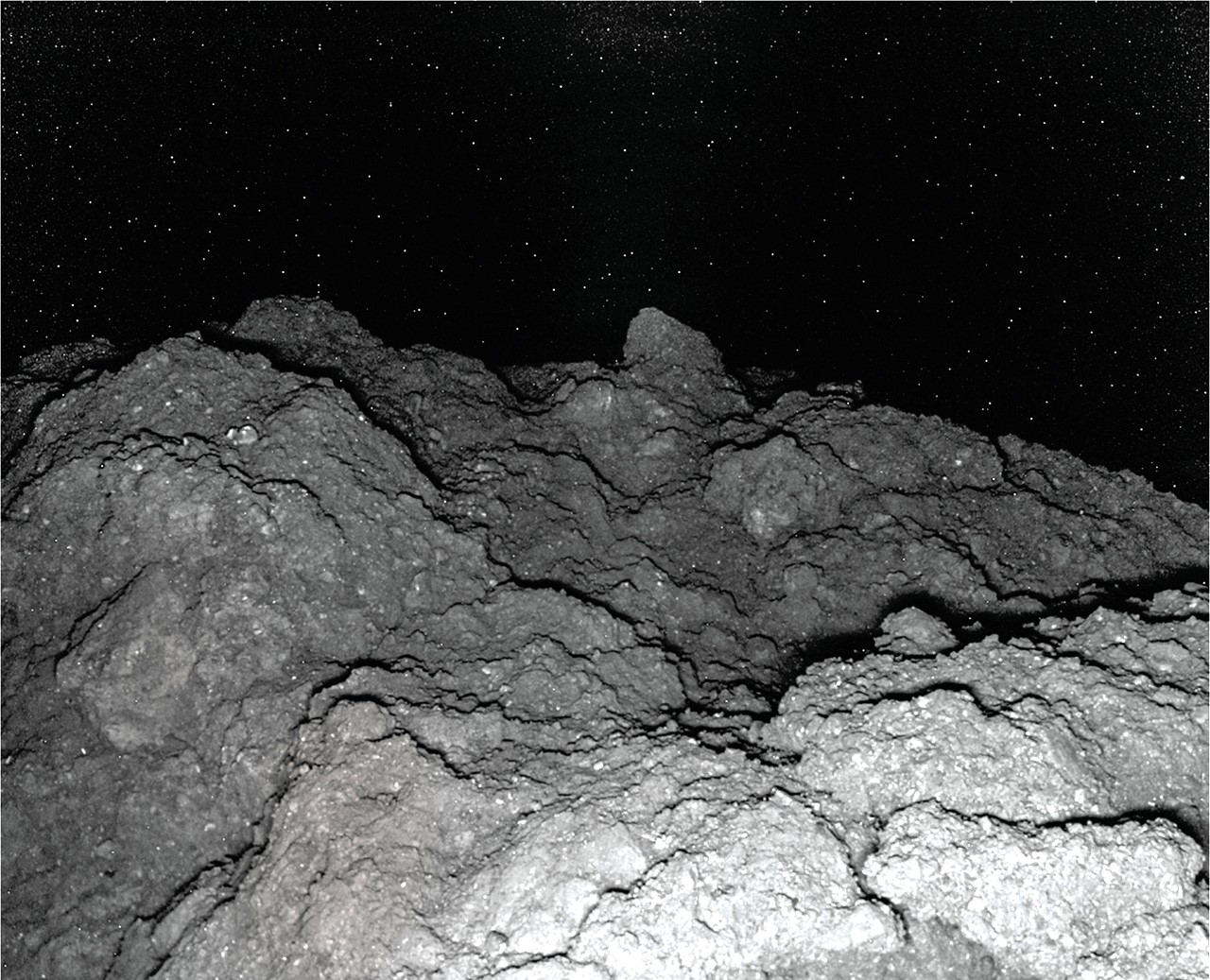}%
            \includegraphics[height=3cm]{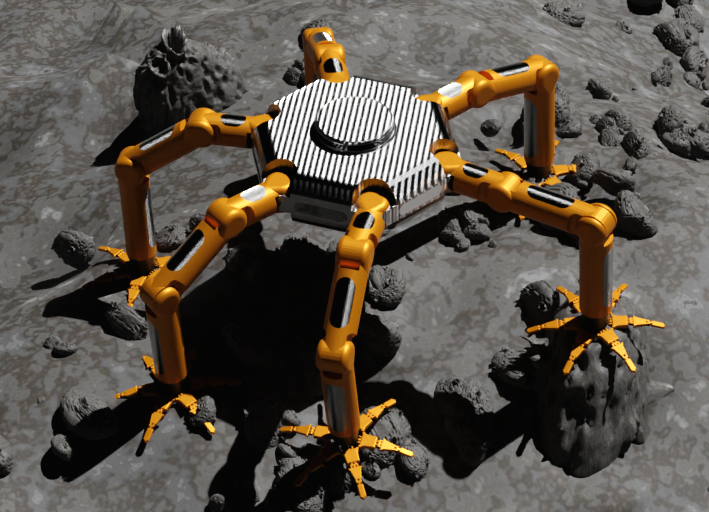}%
        }%
    }
    \setlength{\twosubht}{\ht\twosubbox}
    \centering
    \subcaptionbox{Surface of (162173) Ryugu\label{fig:ryugu}}{%
        \includegraphics[height=\twosubht]{images/ryugu.jpg}%
    }
    \quad \quad
    \subcaptionbox{SCAR-E concept rendering\label{fig:SCAR-E}}{%
        \includegraphics[height=\twosubht]{images/SCAR-E.png}%
    }
    \caption{Asteroid mining prospect for the SCAR-E robot}
\end{figure}

The presented gripping system has been developed to equip 1R~\cite{fsi} class legged robots such as SCAR-E (Fig.~\ref{fig:SCAR-E}). SCAR-E is an omnidirectional hexapod robot prototype designed for terrestrial and space applications. It has been designed to perform horizontal and vertical locomotion in uneven environments. The total span of the robot is \SI{1.6}{\metre} for \SI{20}{\kilogram}. The end effector of each leg can be interchanged between a simple foot for basic walking, a gripping system for grasping and climbing or a tool adapted to the specific task at hand. These tools can be combined to create the optimal configuration for the task.

\subsection{Gripping system for space robots}
To overcome the disadvantages of wheeled rovers, legged robots designed for space exploration need to be equipped with efficient grippers that will allow them to climb extreme terrains. The main targets for the use of legged robots in space exploration are rocky terrains such as lava tubes on the Moon~\cite{lunarcave} or the surface of asteroids. In both cases, the gripper must be reliable and provide a stable anchor point to the ground. To do so, microspines were used to hook the asperities of rocky targets and concentric tangential forces were applied to secure the gripper to the target. Several microspine gripping mechanisms have been developed, such as Parness's LEMUR-II gripper~\cite{jplgripper}, Nagaoka's passive spine gripper \cite{freegrip}, or Li's microspine gripper~\cite{microspinegrip}. Although the geometry of the microspine offers some degree of compliance with the target shape, the fixed structure of the finger does not provide adaptability to variations in the range of 1 to 10 cm and is limited to rather flat surfaces. This was observed by Newill-Smith et al. who proposed the use of a multiphalanx gripper for asteroid exploration~\cite{multiphalanx}. 
\subsection{Soft-gripping technology}
Soft-gripping is an emerging field of research in the world of robotics. Using compliant or deformable fingers actuated with pneumatic~\cite{pneuma}, tethers~\cite{softgripper, wirejam}, or mechanical systems~\cite{mechact}, a robot can grasp objects of different sizes, shapes, materials, and textures. The goal of soft-gripping is to maximize the contact surface between the fingers and the targets to apply controlled pressure and maintain the grasping action with maximal friction. The friction force applied to the target can also be improved by using adhesives~\cite{adhesive}, or engineering materials similar to the skin of a gecko finger~\cite{gecko}.
Other approaches include the use of vacuum-actuated granular pouches to create perfect match contact with the target object~\cite{granjam, pressurejam}.
Although these technologies show promising results in the efficient grasp of random-shaped targets, they are limited by extreme temperature changes, a harsh and dusty environment, as well as payload and volume constraints\cite{spacegripper}. 
This research proposes a soft gripping solution that uses microspines and compliant fingers to apply tangential forces and pressure on the target to allow climbing and object manipulation in space.
\section{Gripping system}

\begin{figure}[ht]
    \centering
    \begin{subfigure}{0.4\textwidth}
        \includegraphics[width=\textwidth]{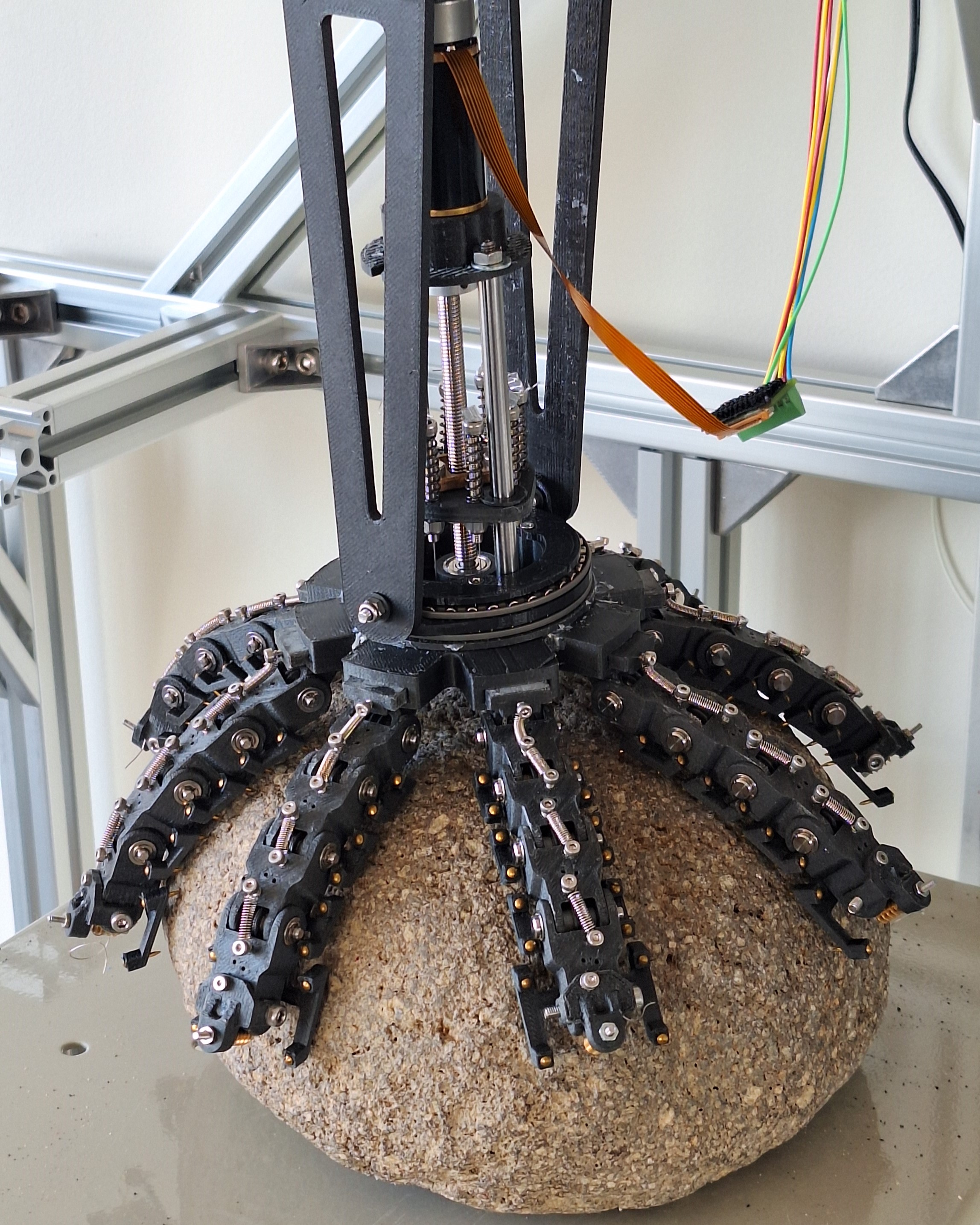}
        \caption{Gripper side view}
        \label{fig:gripper1}
    \end{subfigure}
    \quad \quad
    \begin{subfigure}{0.4\textwidth}
        \includegraphics[width=\textwidth]{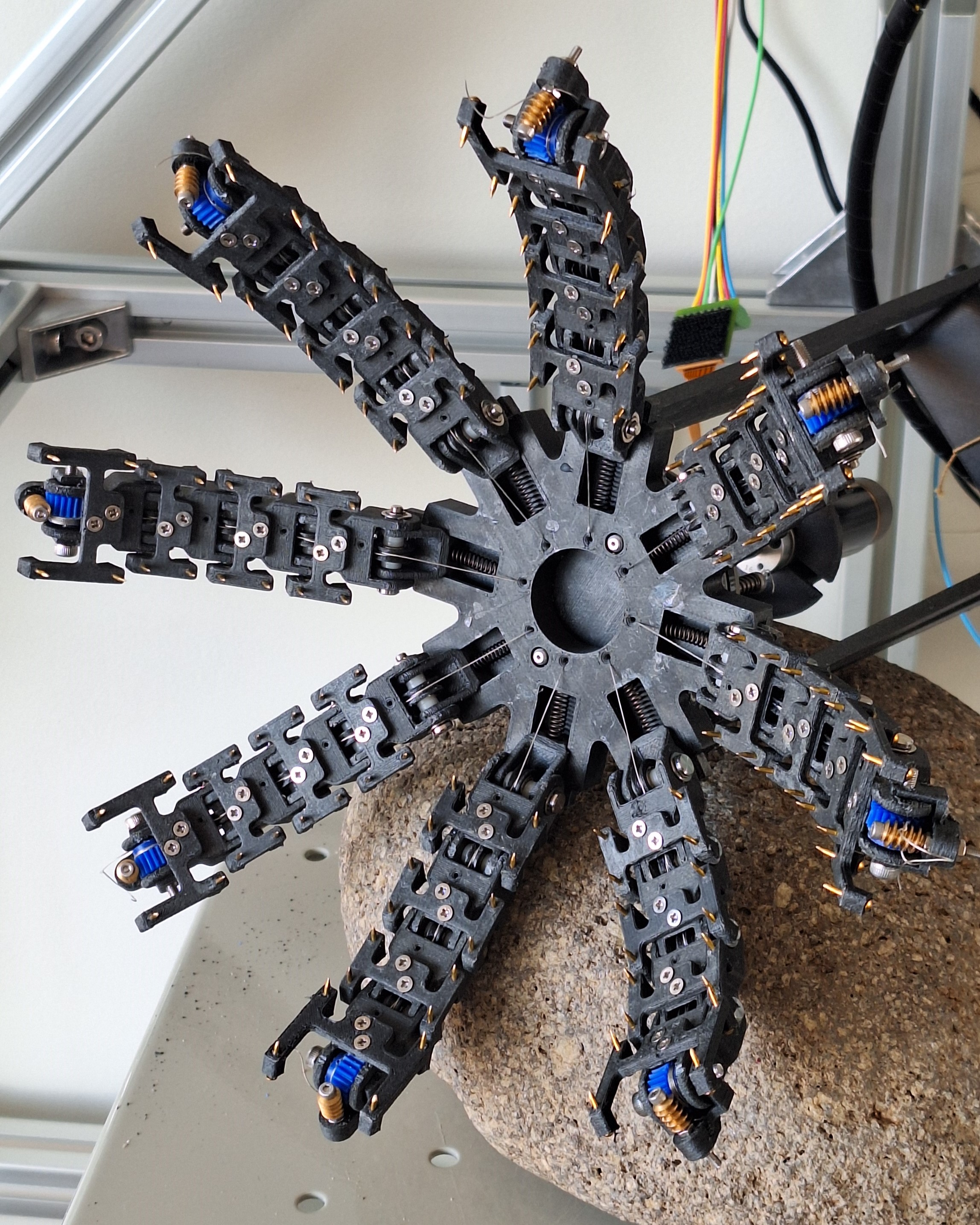}
        \caption{Gripper bottom view}
        \label{fig:gripper2}
    \end{subfigure}
    \caption{Gripper prototype}
    \label{fig:gripproto}
\end{figure}

The main advantage of soft-gripping technologies is the versatility of the fingers to conform to the shape of its designated target while grasping. This maximizes the contact surface and increases the resulting gripping force. In space environments, polymers and engineered materials are subject to changes in properties as a result of extreme temperature variances, and compliance may not be ensured. For this reason, the proposed gripper is designed with purely mechanical compliance and actuation. To further improve the gripping performance, additional degrees of freedom have been added to each finger for better adaptability. This also allows the gripper to apply a tangential force to the surface of the target and increase the hooking action of the spines onto the porous surface of rocks. A gripper prototype (\ref{fig:gripproto}) was built using carbon-fiber infused nylon 3D printed parts and stainless steel spines, for a total mass measured at \SI{0.565}{\kilogram}.

\subsection{Principle}
\label{principle}
\begin{figure}[t]
    \centering
    \includegraphics[width=0.85\columnwidth]{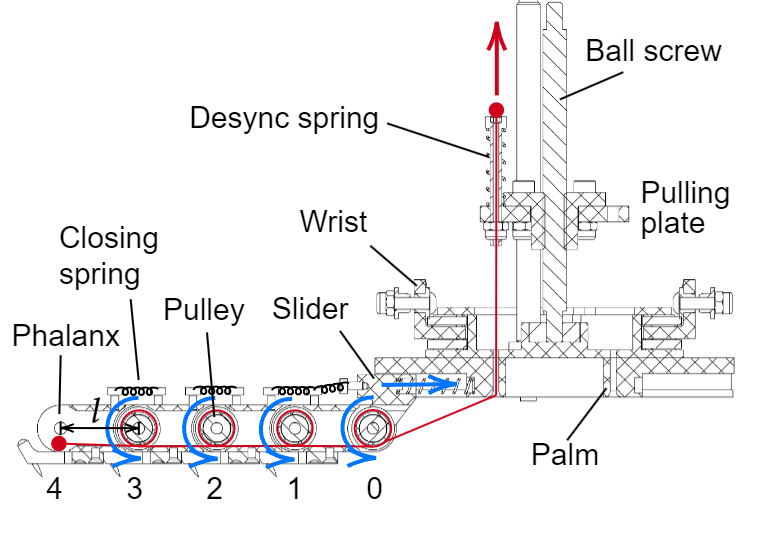}
    \caption{Section view of the gripping system}
    \label{fig:gripper}
\end{figure}

To offer the advantages of soft gripping in space environments, the proposed gripping system relies on multiple tendon-driven fingers divided into 4 phalanges. The tether is attached to the extremity of the finger and is guided through a series of pulleys connected to each joint of the finger, as shown in Fig.~\ref{fig:gripper}. This architecture was previously used in the softgripper developed by Hirose~\cite{softgripper}, which shows great results in matching irregular shapes. Each phalanx is identical in length $l$ and shape to provide a modular option for longer or shorter fingers. The aim of Hirose's research was to generate uniform pressure along the contact areas of the fingers by adjusting the diameter ratio between two consecutive pulleys, which allowed a controlled and gentle grasping of the target. In the case of grasping rocky targets, a more power-oriented approach was taken. When grasping a convex-shaped target, the fingertips generate more resistive force against 
the detachment motion, as the friction force orientation is normal to the target surface. When using the same diameter of the pulleys along the whole finger, the pressure applied to the target during grasping is distributed toward the fingertip, increasing the efficiency of the gripper. The pressure distribution can be derived from the equations presented in ~\cite{softgripper}. In the case of a series of identical pulleys, the torque $\tau$ created by the tether tension $T$ is constant for each joints of the finger and can be expressed as 

\begin{equation}
    \tau = r \times T 
\end{equation}
with $r$ the radius of the pulley. 

Considering that the torque $\tau$ in each joint is constant, the pressure $p_{j+1}$ applied by phalanx $j$ can be derived as

\begin{equation}
\label{eq:pressure}
\displaystyle
    p_{j+1} = \frac{\tau}{L_j} 
\end{equation}

where $L_i$ is defined as

\begin{equation}
    L_{j} \ =
    \begin{Bmatrix}
        \sum\limits _{p=0}^{n-1-j}\left(\sum\limits _{m=0}^{p} l_{n-m}\right) l_{n-p} & \ j=0\sim n-1\\0 & j=n
    \end{Bmatrix}
\end{equation}.

When the lengths $l$ of all the phalanx are equal, Eq.\ref{eq:pressure} can be simplified to

\begin{equation}
    \label{eq:simpressure}
    \displaystyle
    \forall j \in [0,n] \ \ \ p_{j+1} = \frac{2\tau}{l^2[(n-j)(n+1-j)]} 
\end{equation}

For the gripper presented in this study, each finger is composed of $n = 4$ phalanges. The pressure distribution for each phalanx can then be expressed as

\begin{equation}
    \displaystyle
    \label{eq:presdistrib}
    p_1 = \frac{\tau}{10 l^2} \quad p_2 = \frac{\tau}{6 l^2} \quad p_3 = \frac{\tau}{3 l^2} \quad p_4 = \frac{\tau}{l^2} \ .
\end{equation}

The pressure is distributed along the finger, with the maximum value at the fingertip. For each phalanx, it is equally divided between the spines in contact with the target. The total pressure increases as the length between each joint $l$ is reduced and the number of joints increases. However, physical limitations such as spine attachment, assembly and maintenance, pulley radius, and overall gripper diameter dictated the dimensions.

\subsection{Actuation}

When designing embedded systems and effectors for mobile robotics, it is important to take into account energy efficiency. Gripping mechanisms may require a continuous supply of current to maintain the grasp on the target. Therefore, the gripper actuation has been designed with a self locking feature utilizing a ball screw with a pitch of 1mm. The ball screw creates the linear motion that pulls the tethers to close the fingers with the help of a 5W DC brushless motor. The high level of reduction provided by the ball screw prevents back drive in the system. As long as the motor does not rotate to relieve tension on the tethers, the pulling plate is locked, and the fingers continue to apply pressure on their target. 

In the softgripper design~\cite{softgripper}, the opening of the gripper was controlled using a secondary motor and tether. In this study, the opening action is passively ensured using tension springs placed between each phalanx. Although this solution adds resistance against the closing action of the gripper, it simplifies the overall architecture to a single actuator and one tether per finger. This allows for a lighter and more compact design. 

The challenge with having a single actuator for multiple tethers is that every finger will be synchronously closing and opening at the same rate. For a linear displacement $\Delta z$ of the ball screw nut, each tether will be pulled by the same displacement $\Delta z$. This goes against the goal of soft gripping, where all fingers must be free to individually adapt to the target shape. To counteract this, a desync spring, shown in Fig.~\ref{fig:gripper} was integrated into the actuation chain. This allows each finger to move independently of the other fingers. To account for each finger $i$ needing their own displacement $\Delta z_i$, each tether is attached to the pulling plate using a hollowed screw and a compression spring. During the closing motion, if finger $i$ is locked in position to target, the hollowed screw will sink into the pulling plate and continue its motion at a depth of $h_i = \Delta z - \Delta z_i$. The tether tension $T_i$ will increase at a rate of $T_i = k_{spring} \times h_i$, but finger $i$ will not prevent the pulling plate from moving further. The grasping motion of the target is completed when the equilibrium between the sum of the tensions of each tether $T_i$ and the targeted motor torque $\tau_{motor}$ is reached. This solution is a simple and effective way for the gripper to adapt to a random shape, although it comes at the cost of small disparities in the pressure applied by each of the fingers onto the target. 


\subsection{Microspines for rock grasping}

\begin{figure}[ht]
    \sbox\twosubbox{%
        \resizebox{\dimexpr.9\textwidth-1em}{!}{%
            \includegraphics[height=3cm]{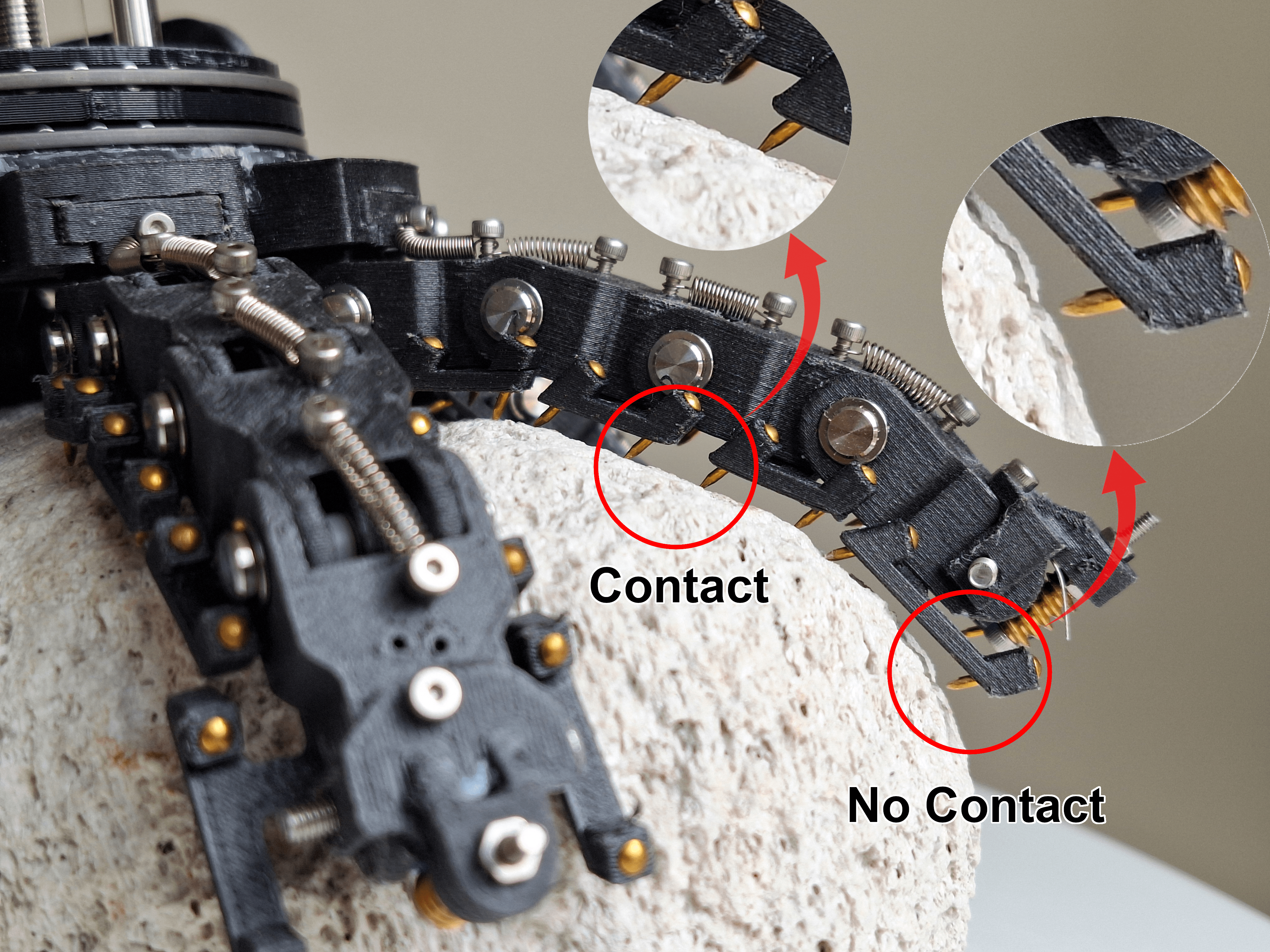}%
            \includegraphics[height=3cm]{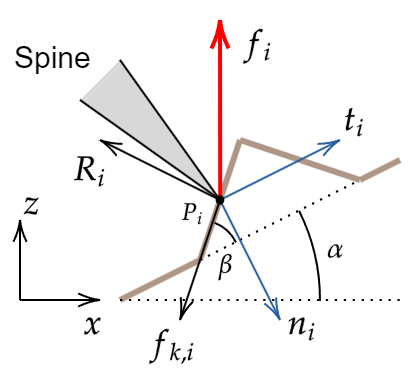}%
        }%
    }
    \setlength{\twosubht}{\ht\twosubbox}
    \centering
    \subcaptionbox{Close-up view of a finger\label{fig:spinecontact}}{%
        \includegraphics[height=\twosubht]{images/spinecontact.png}%
    }
    \quad
    \subcaptionbox{Asperity model\label{fig:spinemodel}}{%
        \includegraphics[height=\twosubht]{images/spinediagram.png}%
    }
    \caption{Spine interaction with a rocky surface}
    \label{fig:spine}
\end{figure}

The use of small sharp spines, referred to as microspines, has been proven to be effective in securing a good grip on the rough surface of rocks~\cite{jplgripper, microspinegrip, microspine, freegrip}. As shown in Fig.~\ref{fig:spine}, microspines latch onto the surface asperities and produce a counteracting friction force $f_{k}$ directed against the detachment motion of the gripper. Following the method presented in ~\cite{freegrip}, such asperities can be modeled as a triangular projection with a slope angle $\beta$, and the local friction coefficient $\mu'$ of a microspine can be expressed as follows.

\begin{equation}
    \label{eq:frictioncoef}
    \mu ^{\prime }=\frac{\mu +\tan \beta }{1-\mu \tan \beta }
\end{equation}

Section~\ref{principle} explained that pulling the control cord creates pressure on the target surface. This pressure, represented locally in Fig.~\ref{fig:spinemodel} as $n_i$, is applied at a single contact point $P_i$ perpendicularly to the local surface of the target, modeled here as a slope of angle $\alpha$. The reaction force $R_i$ can be expressed as:

\begin{equation}
    \label{eq:normalreaction}
    R_i = n_i \cos{\beta}
\end{equation} 

Given the total detachment force $F$, the local detachment force $f_i$ is distributed along the spines that are in contact with the target. The slipping condition can be written as:

\begin{equation}
    \label{eq:slipnormal}
    f_i < f_{k_i}=\mu^\prime R_i \sin{(\alpha+\beta)}
\end{equation}

With only the normal pressure $n_i$, the friction force $f_{k_i}$ is sensible to the slope angles $\alpha$ and $\beta$, and does not guarantee that the microspine will connect with favorably matching asperities. For this reason, a tangential force acting on the base of the finger has been implemented. The base of each finger is linked to the palm via a sliding part as shown in Fig.~\ref{fig:gripper}, with its degree of freedom along the radial axis of the gripper. Equation \ref{eq:normalreaction} can be rewritten as follows. 

\begin{equation}
    \label{eq:reaction}
    R_i = n_i \cos{\beta} + t_i \sin(\beta)
\end{equation} 

The total friction force generated by a spine $i$ of a phalanx $j$ for a finger $a$ with $n_a$ spines in contact with a convex target can be expressed as follows.

\begin{equation}
    \label{eq:slip}
    f_{k_i} = \mu^\prime \left(\frac{r\times T}{L_j}\cos{\beta} + \frac{T}{n_a}\sin{\beta}\right) \sin{(\alpha+\beta)}
\end{equation}

With $r$ the radius of the pulley and $T$ the tension of the finger tether.

\section{Performance evaluation}
A novel gripping system has been introduced that provides the benefits of soft gripping for extreme environments while including microspines technology for gripping rocks in space. The performances of the gripper have been assessed through different experiments conducted on a testing bench and will be presented in this section. 

\subsection{Test environment}

\begin{figure}[ht]
    \centering
    \includegraphics[width=0.8\columnwidth]{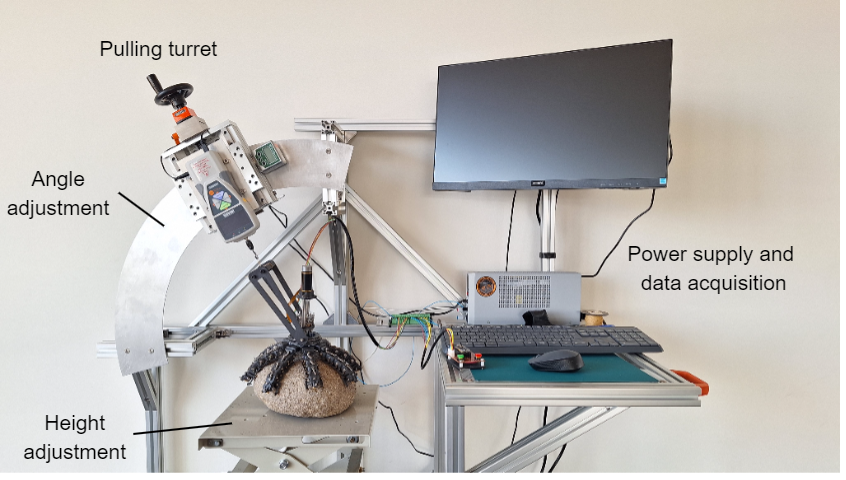}
    \caption{Gripper grasping a basalt rock on the testing bench}
    \label{fig:bench}
\end{figure}

Gripping performance can be characterized by the amount of force that the gripper can withstand before detaching from the target. In the case of climbing robots, the orientation of the detachment force varies greatly depending on the inclination of the slope or cliff being climbed. The force can be a constant load over a long period of time or can be impulse shaped due to the motion of the robot. For the purpose of assessing the gripping performances under a wide range of parameters, a testing bench, presented in Fig.~\ref{fig:bench} was developed. 

The testing bench is composed of a manually actuated pulling turret on which is attached a force meter with a resolution of \SI{0.01}{\newton} and a acquisition frequency of \SI{0.1}{\second}. The turret is mounted on a curved rail to adjust the angle of the pulling action from a range of \ang{0} to \ang{90} that can be dialed with a \ang{0.1} precision using an absolute digital inclinometer. The axial displacement of the turret is measured through a graduated analog display with a resolution of \SI{0.01}{\milli\metre}. The target is bolted to a height-adjustable platform used to calibrate the gripper wrist joint to align with the center of the radius of the curved rail. The wrist of the gripper is a universal joint with roll and yaw freedom, allowing the detachment force to be applied at the center of the gripper regardless of the force orientation. 

\subsection{Gripping performances experiments}

\begin{figure}[t]
    \centering
    \begin{subfigure}{0.32\textwidth}
        \includegraphics[width=\textwidth]{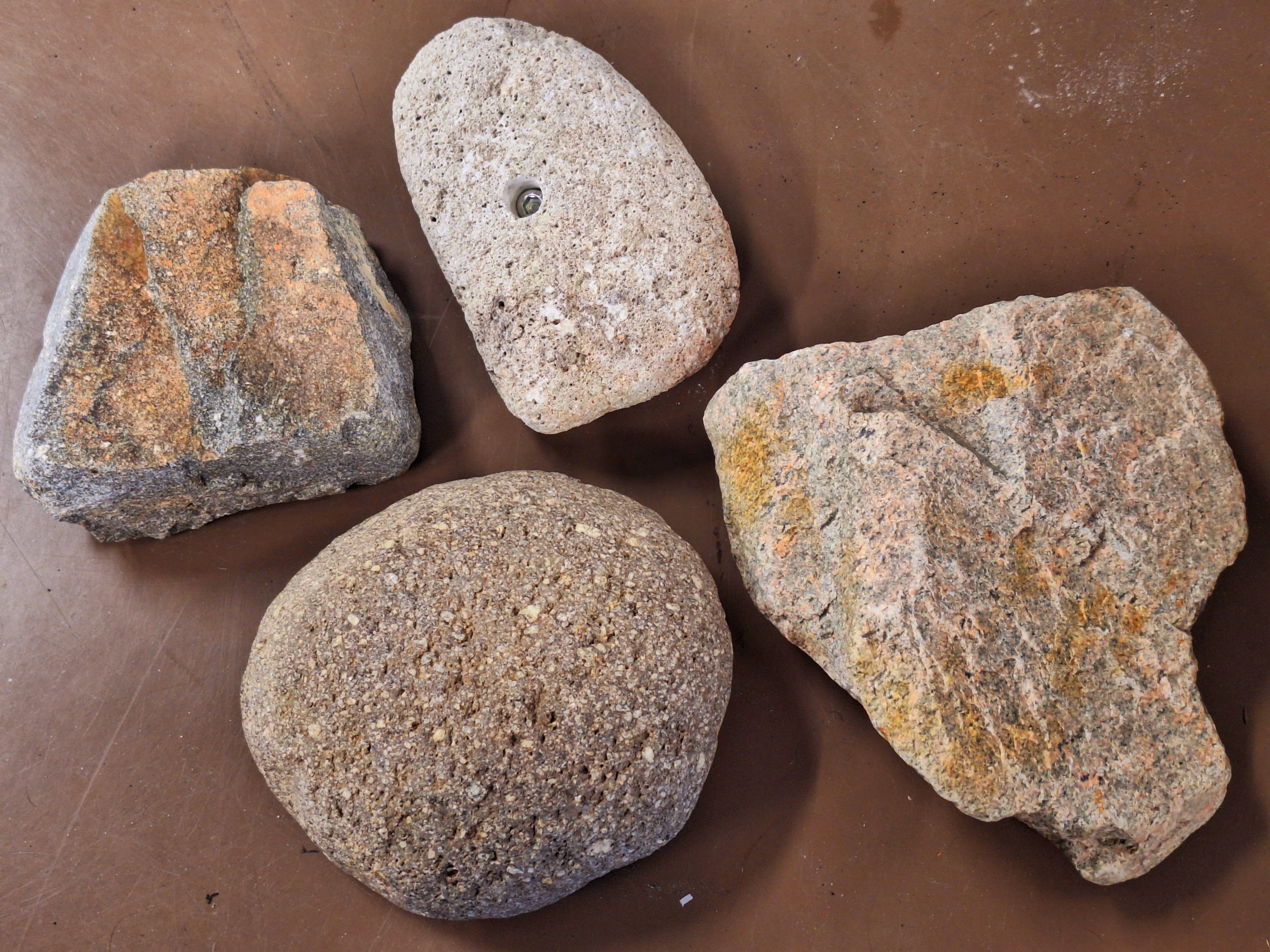}
        \caption{Random targets}
        \label{fig:rocks}
    \end{subfigure}
    \hfill
    \begin{subfigure}{0.32\textwidth}
        \includegraphics[width=\textwidth]{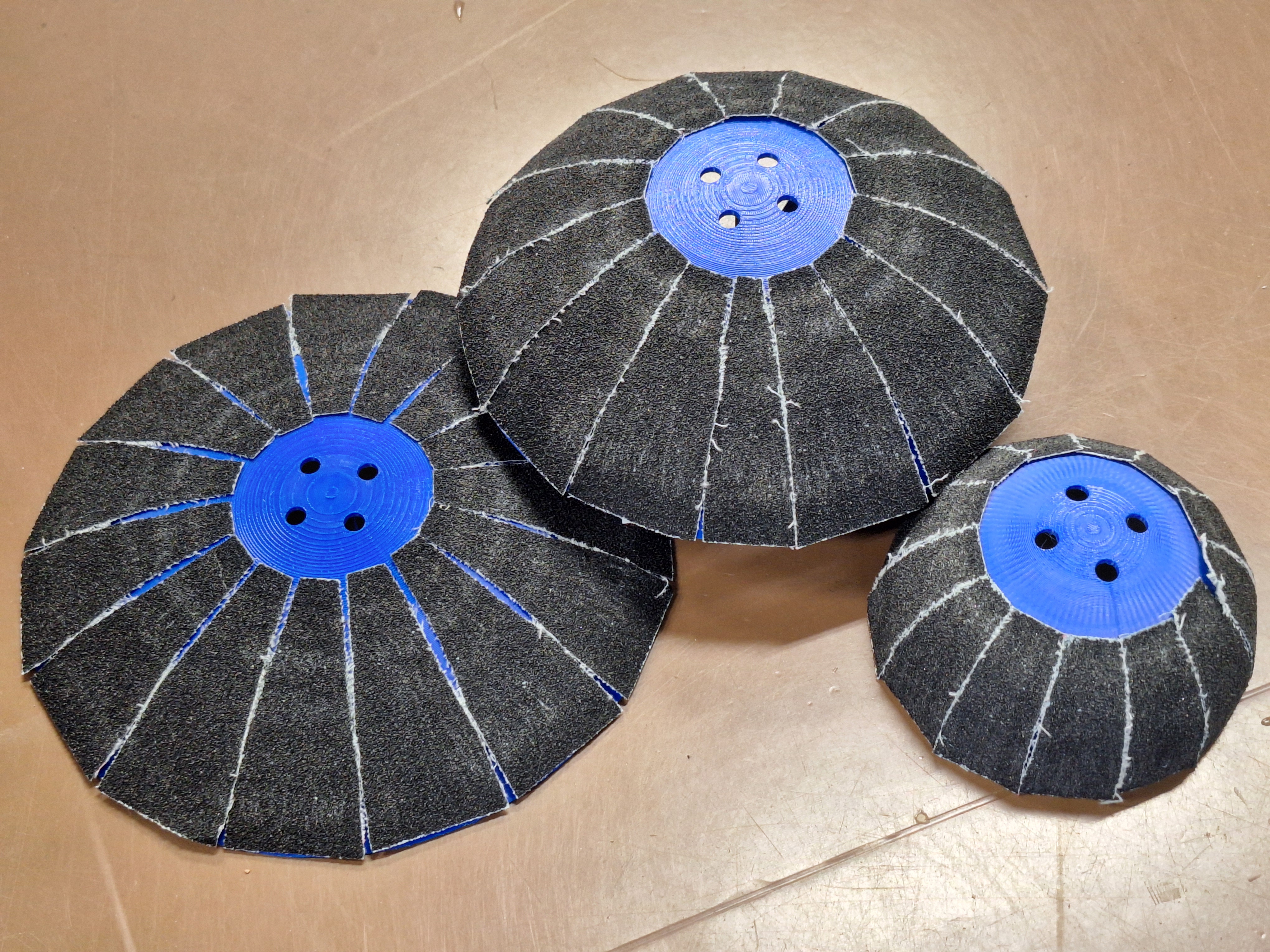}
        \caption{Spherical targets}
        \label{fig:targets}
    \end{subfigure}
    \hfill
    \begin{subfigure}{0.32\textwidth}
        \includegraphics[width=\textwidth]{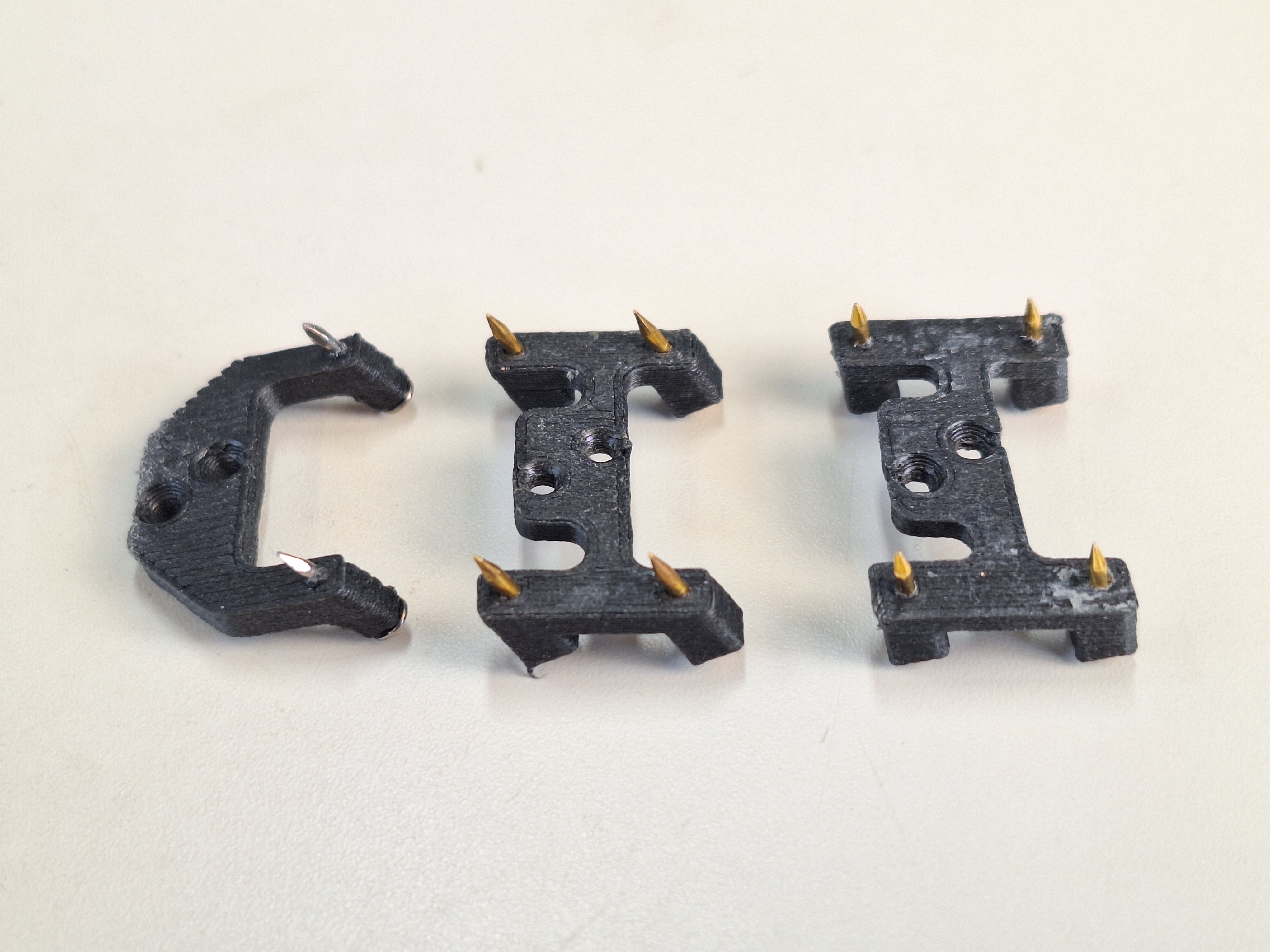}
        \caption{Spine interfaces}
        \label{fig:spineinterface}
    \end{subfigure}
    \caption{Targets and spine interfaces used for the experiment}
\end{figure}

The gripper was tested using multiple random-shaped rocks of different sizes (Fig.\ref{fig:rocks}), but the unevenness of the shape hinders the repeatable aspect of the tests. In order to observe the gripper performances in a wide but controllable range, experiments were conducted on a set of spherical targets covered in \#40 grit sandpaper (Fig.~\ref{fig:targets}). The curvature diameters of the targets were derived from the diameter of the gripper $D=\SI{270}{\milli\metre}$ as follows.

\begin{equation}
   \centering
   \label{targetdiam}
   D_1 = \frac{D}{2} = 135mm
   \quad
   D_2 = D = 270mm
   \quad
   D_3 = \frac{3D}{2} = 405mm
\end{equation}

To assess the impact of the configuration and geometry of the microspines, three types of spine interface (Fig.~\ref{fig:spineinterface}) were designed and tested. The number of spines varies from two to four spines per module with an inclination of \ang{15} or \ang{30}. 

Performances were evaluated by measuring the maximum force required to detach the gripper from its target. The force was applied using the pulling turret and measured with the force meter. This experiment was conducted from a pulling angle range of \ang{0} to \ang{90} with increments of \ang{10} on each target, with each spine configuration repeated 5 times.

A complementary study on gripper behavior was conducted with an increase in load over time. The motor actuation power was tested in the range of \SI{0.15}{\ampere} to \SI{0.275}{\ampere}, creating, respectively, \SI{84}{\milli\newton\metre} and \SI{179}{\milli\newton\metre} torque on the ball screw, in increments of \SI{0.025}{\ampere}.

\section{Results and discussion}

In this section, the results of the experiments will be showcased. After multiple gripping tests on random-shaped rocks (Fig.~\ref{fig:rocks}), the gripping performances averaged at around \SI{35.68}{\newton} with a standard deviation of \SI{17.33}{\newton}.

\begin{figure}[h]
    \centering
    \begin{subfigure}{0.325\textwidth}
        \includegraphics[width=\textwidth]{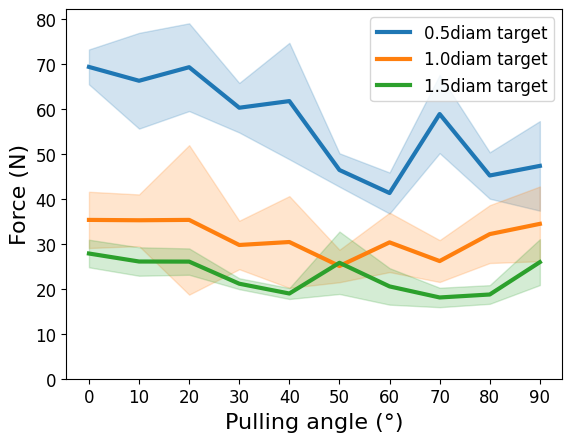}
        \caption{Dual \ang{30} spine}
        \label{fig:gripd30}
    \end{subfigure}
    \begin{subfigure}{0.325\textwidth}
        \includegraphics[width=\textwidth]{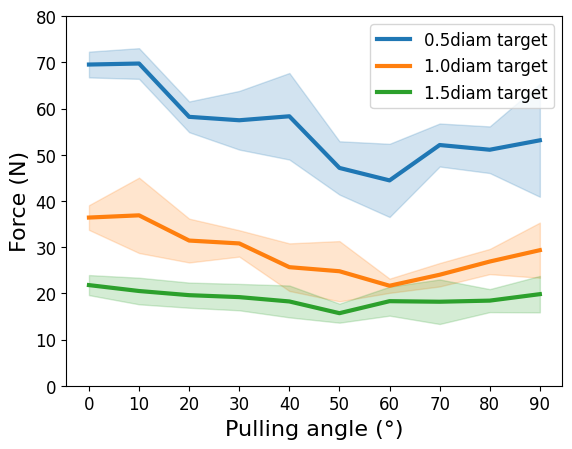}
        \caption{Quad \ang{30} spine}
        \label{fig:gripq30}
    \end{subfigure}
    \begin{subfigure}{0.325\textwidth}
        \includegraphics[width=\textwidth]{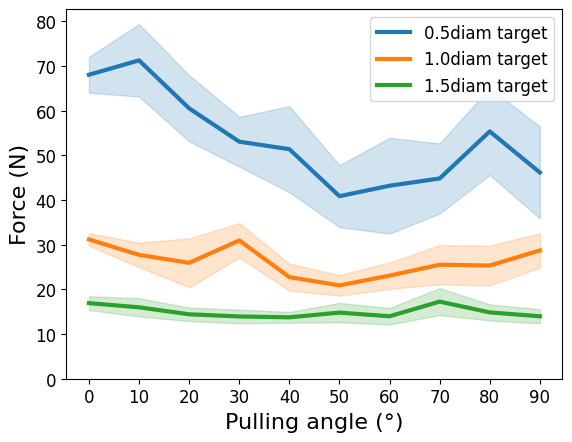}
        \caption{Quad \ang{15} spine}
        \label{fig:gripq15}
    \end{subfigure}
    \caption{Target curvature impact on performances with standard deviation}
    \label{fig:griptarget}
\end{figure}

\begin{figure}[t]
    \centering
    \begin{subfigure}{0.325\textwidth}
        \includegraphics[width=\textwidth]{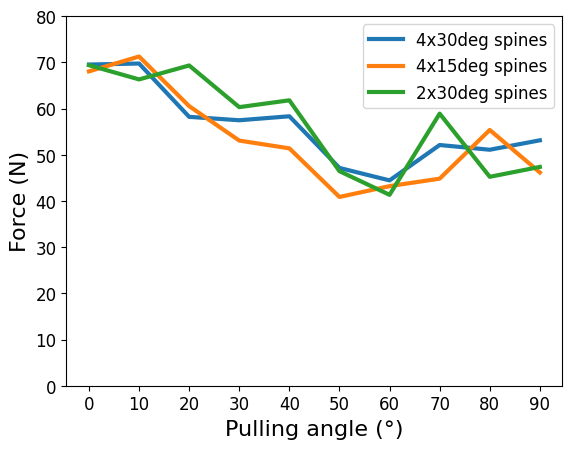}
        \caption{$D_1$ target}
        \label{fig:grip05}
    \end{subfigure}
    \begin{subfigure}{0.325\textwidth}
        \includegraphics[width=\textwidth]{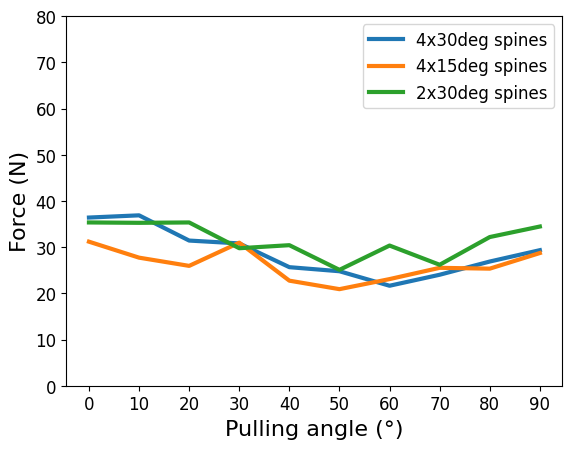}
        \caption{$D_2$ target}
        \label{fig:grip10}
    \end{subfigure}
    \begin{subfigure}{0.325\textwidth}
        \includegraphics[width=\textwidth]{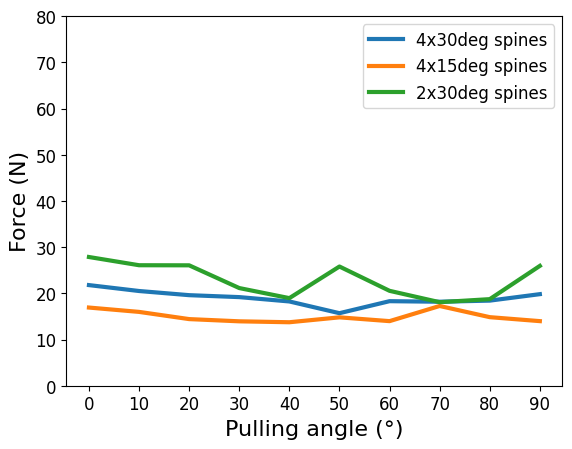}
        \caption{$D_3$ target}
        \label{fig:grip15}
    \end{subfigure}
    \caption{Spine configuration impact on gripping performances}
    \label{fig:gripspine}
\end{figure}

In the results shown in Figs.~\ref{fig:griptarget} and \ref{fig:gripspine}, the variation in gripping performance can be observed. As expected, the gripping force increases as the target diameter decreases as a result of the orientation of the resistive friction force. As the pulling angle increases from \ang{0}, a decrease in performance is observed, followed by an increase around the \ang{80} mark. This behavior can be attributed to how each finger contributes against the detachment motion. At \ang{0}, all fingers contribute equally to oppose the pulling force. As the angle increases, the strain shifts to the opposite side of the detachment, resulting in the opposite fingers carrying more load than the others. This causes imbalance in the gripping action and increases the chances of slipping. As the angle approaches \ang{90}, the opposite side fingers generate a strong resistive force in complete opposition to the detachment motion, therefore increasing the performances. 

In Fig.~\ref{fig:gripspine}, the impact of spine configuration on gripping performance is presented. The results show that while all configurations follow the same tendency as explained in the previous paragraph, a \ang{30} inclination spine fares better overall. This behavior is attributed to the difficulty of the lower-angle spine to secure itself onto a suitable asperity before or after slipping. The number of spines also shows variations in performance. The dual spine configuration seems to hold the highest maximum detachment force, although the standard deviation in Fig.~\ref{fig:griptarget} show a slightly larger variance than the quad spine configurations. The distribution of the finger phalanx pressure onto fewer contact points helps the finger to match the shape of the target and apply a stronger local force, but may be more vulnerable to slipping or poor asperity contact. 

\begin{figure}[ht]
    \centering
    \begin{subfigure}{\textwidth}
        \includegraphics[width=\textwidth]{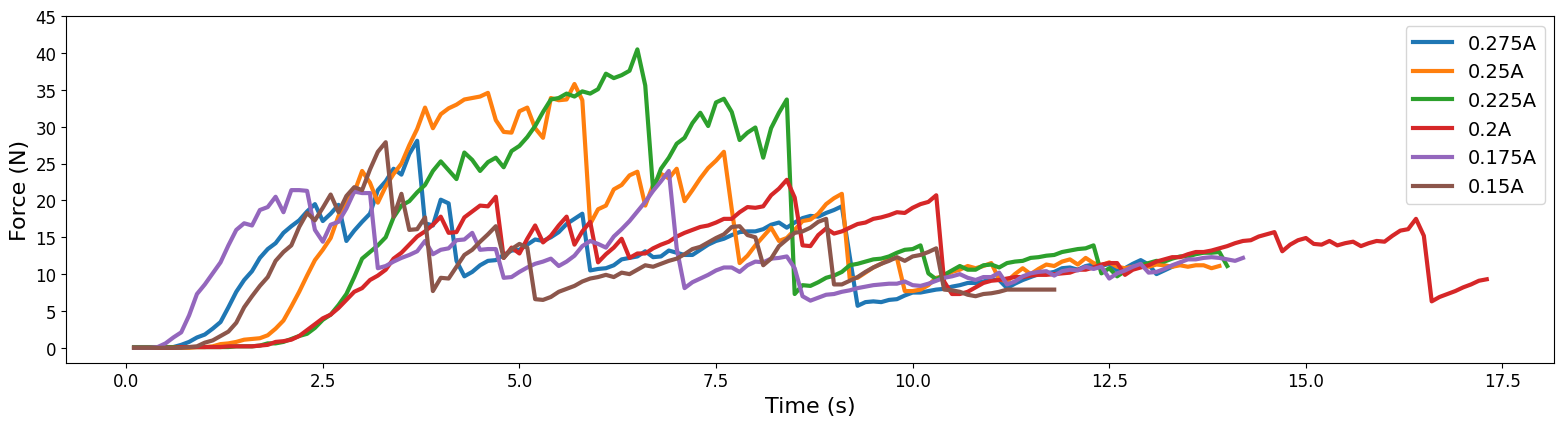}
        \label{fig:gripslip}
    \end{subfigure}
    \begin{subfigure}{0.495\textwidth}
        \includegraphics[width=\textwidth]{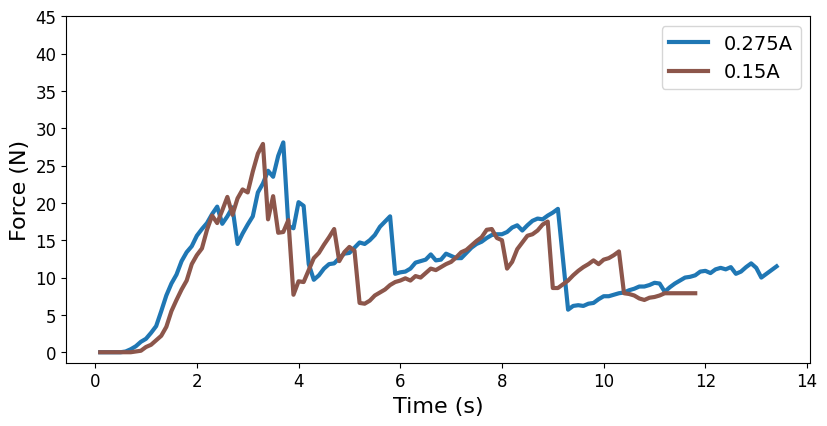}
        \caption{}
        \label{fig:sliplow}
    \end{subfigure}
    \begin{subfigure}{0.495\textwidth}
        \includegraphics[width=\textwidth]{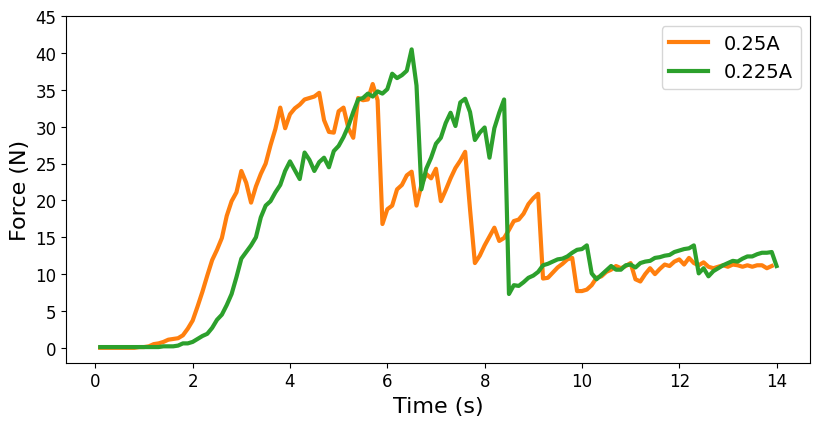}
        \caption{}
        \label{fig:sliphigh}
    \end{subfigure}
    \caption{Gripping performances over time with different actuator current}
    \label{fig:gripcurrent}
\end{figure}

Finally, experiments were conducted on the gripper behavior under increasing load and on the effect of motor torque on gripping performance. The results in Fig.\ref{fig:gripcurrent} demonstrate the slipping action that occurs during gripping. A linear continuous increase in gripping force can be observed until around \SI{20}{\newton}, prior to which the gripper provided a secure and immobile anchor on the target. Then, slipping begins to occur as shown by the sharp drop in gripping force. The first slip does not detach the gripper from the target, but weakens the maximum performances in some cases. In other cases, the slip allows fingers with underperforming contacts to attach to a more suitable area, increasing the maximum performance beyond the first slip limit. The ability of the gripper to recover from slipping mainly lies in the tether tension and torque applied by the actuator. As shown in Fig.~\ref{fig:sliplow}, not enough tension or too much tension can hinder the ability of the slipping spines to latch onto another asperity. For the current gripper prototype, an optimal actuation power between \SI{0.225}{\ampere} and \SI{0.25}{\ampere} generates the best gripping performance (Fig.\ref{fig:sliphigh}).  

In the case of a \SI{20}{\kilogram} legged robot such as SCAR-E, to climb on the ceiling of a lunar cave using a tripod gait, the maximum force required can be estimated at around \SI{10.8}{\newton} per gripper. On Mars, the force required for a similar task is around \SI{24.73}{\newton}. Looking at the experimental results, the gripper is able to provide a reliable gripping action in this range and can be considered a suitable system for climbing in lunar, martian or microgravity environment. 

Further tests will be conducted on the gripper, including the measure of the pressure distribution at each phalanx. The spine interface will be improved to provide more compliance with uneven surfaces and better adherence to the target surface.

\section{Conclusion}

Despite technological advances in prehension technologies for robotic systems, applications in space remain challenging due to the harshness of the environment. This studies aims to develop a novel gripping system for space robots to allow grasping and climbing on rocky terrain such as asteroid surfaces or Moon caves. The gripper aims to combine soft gripping technology to optimize the contact with the target and microspine gripping to latch on rough rocky surface. Its design revolves around the creation of rotation and translation motion of compliant fingers with a single actuator to provide adaptability to the target shape and strong adhesion to its surface. 

The gripping performance was evaluated using multiple parameters, such as the pulling angle, the target shape, the spine configuration, and the actuation power. The gripper is able to provide a reliable anchor for 1R space exploration legged robot such as SCAR-E in reduced gravity environment.


%
%
\DeclareFieldFormat{labelnumberwidth}{\mkbibbold{#1\adddot}}
\printbibliography
\end{document}